# An intelligent satellite must do nothing well

Fabrizio Pilone[1], Giordana Bucchioni[2]

[1]*Dept. of Information Engineering, University of Pisa, Pisa, Italy, fabrizio.pilone@phd.unipi.it, 0009-0002-0013-2575*
[2]*Dept. of Information Engineering, University of Pisa, Pisa, Italy, giordana.bucchioni@unipi.it,* 0000-0002-5654-0730

***Abstract*—Autonomous inspection of a resident space object is usually posed as a contest between classical next-best-view planning and reinforcement learning. We argue that the choice of algorithm is the least informative of the design axes. An inspector spacecraft has no rest state: when it stops thrusting, orbital dynamics carry it along a natural trajectory that can be used to observe the target at no fuel cost. On a controlled ladder in which one design axis is varied at a time, fuel efficiency turns out to be set by three things: whether the agent is allowed to drift, whether its state records where drifting has taken it, and whether its planning horizon is long enough for the benefit to appear. Removing any one of them doubles the fuel bill or more. The algorithm still matters, but through what it can represent: a model-based agent whose learned world model ignores fuel spends several times more of it. Finally, no learned policy is safe by construction, and a control barrier function filter beneath the decision layer removed every keep-out-zone violation in two thousand evaluation episodes.**

## I. Introduction

Close-proximity inspection is a prerequisite for servicing, life extension and removal of resident space objects: an inspector spacecraft must image the whole surface of a passive, possibly damaged target with little support from the ground [1]. The inspector satellite usually cannot be teleoperated because of the presence of communication delays and possible disturbances. For this reason, it is necessary for the satellite to be enhanced with intelligence features for autonomous task execution.

The inspection problem is usually framed as a choice of algorithm between classical Next-Best-View (NBV) and Reinforcement Learning (RL) [2], however this is only one of the angles from which to look at intelligence: the best algorithm cannot make up for sloppy representation of the satellite's actions, and perfect modelling cannot supersede safety assurance. For this reason, an "intelligent" satellite not only has to be good at its job, but also be flexible, autonomous and safe. We carried out a controlled ladder, on which NBV planners and RL-enhanced agents are evaluated under identical conditions, varying one design axis at a time: information availability, state representation, algorithm choice and action space all showed their impact on mission effectiveness and efficiency.

## II. Problem Setting

The environment where an orbital inspector lives is very different from that of most robots. Its motion relative to the target object is governed by the Clohessy-Wiltshire (CW) equation [3]:

$$\begin{cases} \ddot{x} = 3n^2x + 2n\dot{y} \\ \ddot{y} = -2n\dot{x} \qquad + u \\ \ddot{z} = -n^2 z \end{cases}$$

Where n is the orbital rate of the target body. This motion has no rest state, meaning that the null action does not hold it in place, but lets it drift along a natural trajectory: drift is not a disturbance, but a motion primitive. Learning to exploit this motion for fuel-free trajectories places demands on the agent: it must be allowed to do nothing, it must understand where doing nothing has taken it, and it must value a benefit appearing over the long term. At the same time, fuel cannot be replenished, and entering the Keep Out Zone (KOZ) around the target is unrecoverable, so the real point is that average behaviour cannot certify safety. We therefore ask where the intelligence of such a robot resides: in the algorithm, the state representation, the action set, the planning horizon, or a control-theoretic layer that contains it.

## III. Case Study

This study considers a sphere of 50 m radius, with a 10 m KOZ clearance moving on circular orbit of 400 Km; the inspector's camera has a 30° half-angle field of view and sees only sunlit surface [4].

The investigation started with a *viewpoint selection* problem, where the agent chose among 30 viewpoints on fuel-free, 2:1 projected circular orbits (PCO) with bi-impulsive transfer trajectories, or the action of passively drifting under the free-floating dynamics. The second mode considered a *thrust level* RL agent, commanding ΔV every 10 seconds under a dense coverage reward, minus a fuel penalty and a collision termination policy (as in [2]), and trained with Proximal Policy Optimization [5] and DreamerV3 [6].

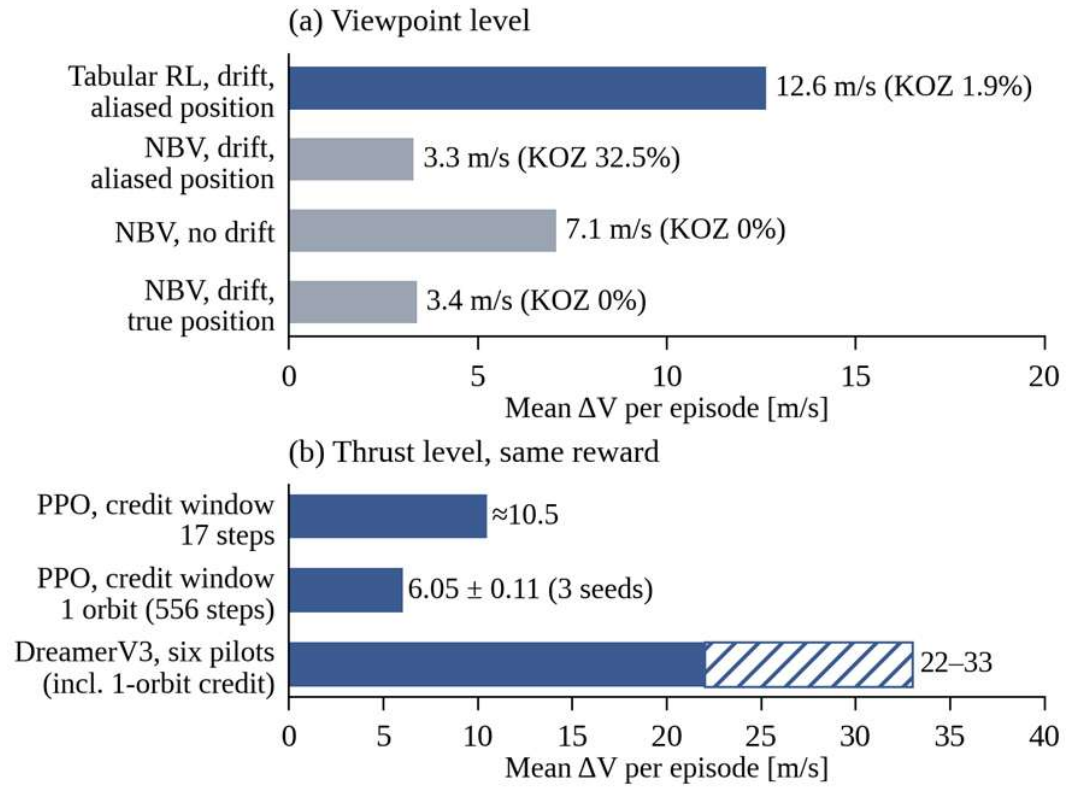

*Fig. 1: Fuel spent per episode. (a) Viewpoint level [4]: all agents share the sector-level state; KOZ = share of episodes with a keep-out-zone entry. (b) Thrust level: continuous ΔV every 10 s.*

## IV. Results

The ablation ladder performed on the case study allowed to identify the main levers acting on fuel efficiency without sacrificing mission success. Since every algorithm and variation reached 98% of target coverage, fuel and safety are then used to discriminate the effects.

### *A. Intelligence is in what the agent can do and represent*

Two capabilities set the fuel bill at viewpoint level: the action space and the state representation. Without the drift action the fuel-aware planner pays for every repositioning, 6.8 m/s; allowing it cuts fuel to 3.4 m/s, with 63% of decisions spent drifting, the single largest fuel-saving lever [4].

At the same time the agent has to know where drifting has taken it: on a discrete viewpoint grid, the drift action leaves the satellite between grid nodes, in a state the encoding cannot represent. A planner given that aliased state still believes to sit at the last commanded viewpoint, and plans its transfer from the wrong position, resulting in a highly collision-prone action, causing KOZ entries in 32.5% of episodes. The tabular agent, penalised for those collisions during training, drew the rational conclusion: since a safe drift and an unsafe one look identical, it learnt not to drift, drifting in 1.3% of decisions and spending 12.6 m/s. Figure 1 shows the main results obtained from the evaluation of different approaches with the effects of various ablations and tests over them, showing that an action is worth nothing if the state cannot show its consequences; whether the robot plans or learns is secondary.

### B. *Intelligence must plan on the body's time scale*

As described in the previous section, fuel efficiency is deeply connected to exploiting the natural dynamics of the system. For this purpose, the agent must be able to learn that actions have long-term advantage: over the next few decisions a manoeuvre always buys coverage sooner, while the fuel saved and the view collected later accumulate over a large fraction of the orbital period. Several reward and action-space investigations left an 8-12 m/s floor: an explicit coast action was chosen in less than 1% of decisions. Extending the credit window [7] to one orbital period roughly halved fuel without impacting mission success (under 6 m/s with 98% coverage over multiple seeds). Offering the right action is necessary but not sufficient: its value must be visible within the agent's planning horizon. Figure 2 reports the 3D trajectories of agents with different planning techniques: fuel consumption is related to the fuel-free drift actions and better pathing.

### C. *The algorithm matters through what it can represent*

A second, model-based technique was then assessed: DreamerV3, which learns a model of the environment and trains on imagined rollouts. Used on the same environment and reward, this algorithm stayed at 22-33 m/s across multiple axs of perturbation, such as reward tuning, and imagination horizon. Probing its world model showed why: cutting ΔV by 72% changed the imagined return by a negligible amount. The fuel term was abstracted away, leaving no room for fuel sensitivity: the internal model can discard what matters physically.

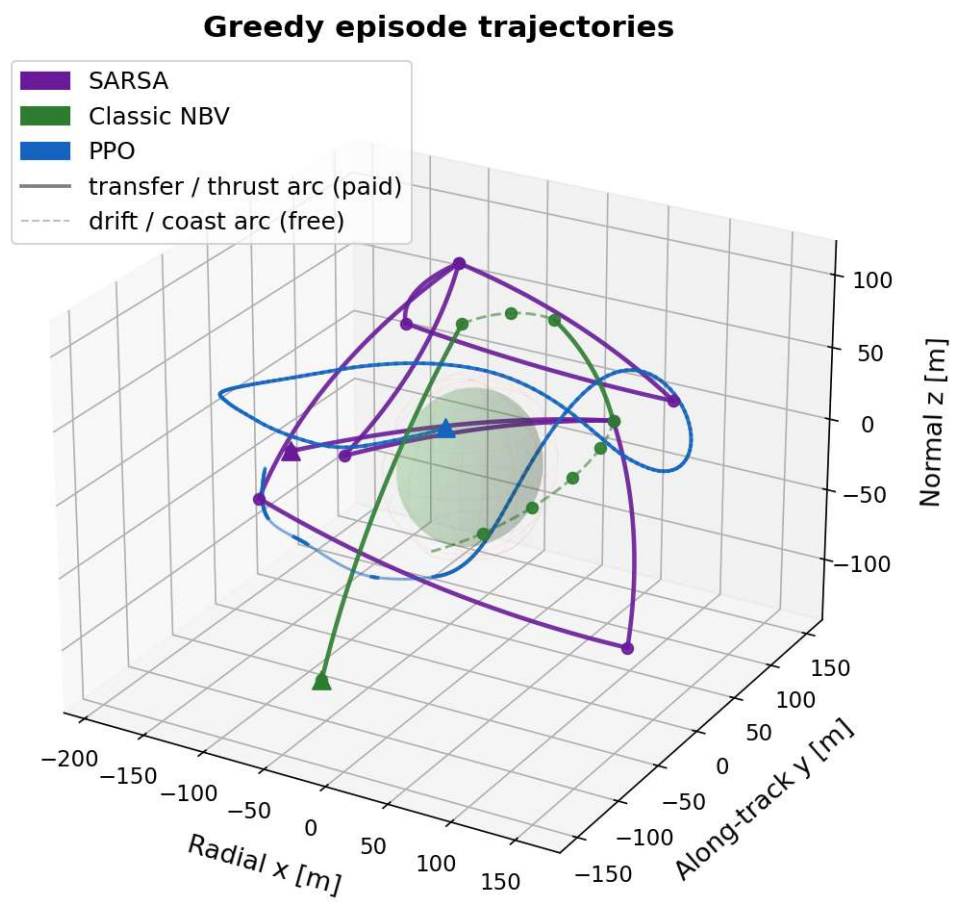

*Fig. 2: 3D trajectories followed by the three different agents (classic, RL and DeepRL). Full lines are arcs flown using fuel; dotted lines are arcs flown exploiting natural dynamics*

### D. *Learning is not sufficient*

Learned policies are reliable in the typical case, not the worst: the PPO agent still fails in around 7% of episodes per seed, and the tabular agent's worst episode covers 55% of the target. In space close-proximity operations safety is always the first requirement, and it cannot be certified by the typical episode. For this reason we placed a Control Barrier Function (CBF) filter based on active set invariance ([8], [9]) beneath the decision layer. CBFs can be seen as a Quadratic Programming problem, with the definition of a safe set $C = \{x | h(x) \geq 0\}$from where the agent must never depart. We defined the set with $h(x) = \sqrt{2a_{max}(||r|| - R_{koz})} + v_p$, representing the outside of the KOZ $(||r|| - R_{koz})$, with manageable closing velocity $(v_p)$ and maximum acceleration authority $(a_{max})$. The impact on the desired control command was minimized, and the safety shield was proved effective on 2000 shielded episodes, where the fraction of KOZ violations collapsed from 33% without the CBF, to 0 with the CBF (95% upper bound 0.15%), with mean coverage of 99.2%.

## CONCLUSIONS

Read against the classical opposition between representational and reactive robots [10], the evidence favours a hybrid. The body contributes free motion, as embodied accounts predict [11], but only for an agent that may exploit the environment, knows where it takes it, and plans far enough ahead to value it. Learning contributes cheap long-horizon behaviour but inherits the blind spots of its state, action set and internal model; a thin reactive layer grounded in control theory contributes the guarantees learning cannot. We therefore adopt a hierarchy: a planned or learned decision layer chooses where to look, low-level skills trained with orbit-matched credit execute the transfers, and a CBF shield enforces invariants underneath.